\documentclass[conference]{IEEEtran}
\usepackage{amsmath,amsfonts,amssymb,amstext,mathtools}
\usepackage{algorithm, algpseudocode}
\usepackage[font=small]{caption}
\usepackage{graphicx}
\usepackage{cite}
\usepackage{fixmetodonotes}
\usepackage{todonotes}
\usepackage{float}
\usepackage{comment}
\usepackage{multirow}
\usepackage{flushend}
\usepackage[colorlinks,citecolor=blue!70,linkcolor=red!80!green!90]{hyperref}

\renewcommand\thesubsection{(\roman{subsection})}
\begin{document}
\raggedbottom
\title{End-to-end learning of brain tissue segmentation from imperfect labeling }

\author{\IEEEauthorblockN{Alex Fedorov\IEEEauthorrefmark{1}\IEEEauthorrefmark{2}, Jeremy Johnson\IEEEauthorrefmark{3}, Eswar Damaraju\IEEEauthorrefmark{1}\IEEEauthorrefmark{2}, Alexei Ozerin\IEEEauthorrefmark{4}, Vince Calhoun\IEEEauthorrefmark{1}\IEEEauthorrefmark{2}, Sergey Plis\IEEEauthorrefmark{1}\IEEEauthorrefmark{2}}
\IEEEauthorblockA{\IEEEauthorrefmark{1}The Mind Research Network, Albuquerque, USA}
\IEEEauthorblockA{\IEEEauthorrefmark{2}ECE Dept., University of New Mexico, Albuquerque, USA}
\IEEEauthorblockA{\IEEEauthorrefmark{3}New College of Florida, Sarasota, USA}
\IEEEauthorblockA{\IEEEauthorrefmark{4}Moscow Institute of Physics and Technology, Russia}}

\maketitle
\begin{abstract}
  Segmenting a structural magnetic resonance imaging (MRI) scan is an important pre-processing step for analytic procedures and subsequent inferences about longitudinal tissue changes.
  Manual segmentation defines the current gold standard in quality but is prohibitively expensive.
  Automatic approaches are computationally intensive, incredibly slow at scale, and error prone due to usually involving many potentially faulty intermediate steps.  In order to streamline the segmentation, we introduce a deep learning model that is based on volumetric dilated convolutions, subsequently reducing both processing time and errors.
  Compared to its competitors, the model has a reduced set of parameters and thus is easier to train and much faster to execute. 
  The contrast in performance between the dilated network and its competitors becomes obvious when both are tested on a large dataset of unprocessed human brain volumes. The dilated network consistently outperforms not only another state-of-the-art deep learning approach, the up convolutional network, but also the ground truth on which it was trained.
  Not only can the incredible speed of our model make large scale analyses much easier but we also believe it has great potential in a clinical setting where, with little to no substantial delay, a patient and provider can go over test results.
\end{abstract}

\section{Introduction}
\label{sec:intro}

Segmentation into tissue types is an important task for analysis of tomographic images of human organs.
One dramatic example is cancer tissue segmentation for treatment and/or surgery planning.
Routine procedures in medicine and research also heavily rely on the process.
For example, segmentation is an inherent part of voxel based morphometry (VBM)~\cite{Ashburner2000805}, a successful neuroimaging technique for analyzing MRI images, and volume/surface analysis and visualization approaches~\cite{dale1999cortical}.
For brain imaging, segmentation of brain into gray and white matter usually follows the extraction of the brain from the MRI image (usually T1).
That may involve several steps, one of which is the stripping of the skull, which itself is a nontrivial problem~\cite{kleesiek2016deep}.
Of the many segmentation algorithms, some of the most popular include SMP~\cite{friston1994statistical} and Freesurfer~\cite{dale1999cortical}.
However, neither are ideal and although Freesurfer is often considered the state of the art it takes a considerable amount of time to process a single T1 volume.\footnote{Note, a typical FreeSurfer run involves more work than just segmentation.}

It is clear that an accurate segmentation is more valuable than a less accurate one as it can lead to better inferences about the brain and disease.
The speed of processing may not seem as important if the processing time of existing tools is acceptable (e.g. within tens of minutes).
However, with the growing importance of large scale studies and scanning projects, which deal with volumes numbered in the thousands, any decrease in the processing time is important and can have an impact on big data oriented medical research.
Furthermore, a simple, fast, and accurate segmentation procedure may turn many current batch processes into interactive explorations, providing significant benefit for neurologists, radiologists, and in the end the patient, when used in medical settings.

In pattern recognition and image segmentation we traditionally rely on human experts for the ground truth. 
Segmenting MRI images (and other medical images as well) is a difficult task, as the expert has to operate in 3D with large (on the order of tens of thousands of points (voxels)) datasets. 
In order to classify a volume using a 2D display, an expert has to label the volumes working a slice at a time, which could lead to inconsistencies.
Different filtering techniques, different scan sequences, and different modalities (e.g. diffusion tensor imaging (DTI)), when simultaneously taken into consideration with the original T1 MRI scan may provide additional information able to improve the quality of segmentation.
However, for a human expert simultaneous consideration of this information is a tedious and impractical task and thus the advantage of multimodal labeling is often necessarily ignored.
Due to these challenges, not only the speed of processing but also the accuracy may suffer.

End-to-end training via deep learning has had a number of successes in highly relevant tasks of pattern recognition and classification.
The approach, however, requires substantial amounts of training data (mostly due to the great flexibility of the models), which is, as we have noted, difficult to obtain if we strive for human annotated training volumes.
Fortunately, the field of brain imaging has developed reliable tools, that can produce automatic segmentations of brain tissue.
Automatic processes are much more scalable than ones involving human experts and the field has accumulated a substantial collection of training data which is automatically labeled.
For example, the Human Connectome Project (HCP)~\cite{van2013wu} provides a FreeSurfer segmentation for each of its subjects.
Although FreeSurfer is indeed a state of the art tool for the task, and it arguably produces very reliable segmentations~\cite{eggert2012accuracy}, its accuracy is unstable.\footnote{Although can be improved with human expert intervention.}

In this paper we generalize dilated convolutional networks~\cite{yu2015multi} to the volumetric case to obtain a model for volumetric segmentation, which we call MeshNet.
We will see that MeshNet has a significantly reduced number of parameters compared to other deep learning models used for the task, in turn reducing the amount of training data required for the model to generalize and provide high accuracy segmentations on testing data.
Importantly, parameter reduction leads to faster model than the state of the art competitors.
Another important property of this model is its ability to provide correct improvised predictions even when trained on imperfect data (we compare it to mistakes that FreeSurfer makes).

\section{Methods}
\label{sec:methods}
Although, a great number of algorithms and implementations exists for MRI segmentation (a partial list is available at the rankings page of \href{http://mrbrains13.isi.uu.nl/results.php}{MRBRainS Challenge} at \href{http://mrbrains13.isi.uu.nl/results.php}{http://mrbrains13.isi.uu.nl/results.php}) many of the highest performing ones are based on volumetric convolutional neural networks (CNNs). As a competing model we use U-Net~\cite{cciccek20163d,ronneberger2015u}.

Basic equation for discrete volumetric convolution is expressed as
\begin{equation}
(k*f)_{(x,y,z)} = \sum_{\bar{x}=-a}^a \sum_{\bar{y}=-b}^b \sum_{\bar{z}=-c}^c k(\bar{x},\bar{y},\bar{z})f(x-\bar{x},y-\bar{y},z-\bar{z}),
\label{eq:3dconv}
\end{equation}
where $a$, $b$, $c$ are kernel bounds on $x$, $y$ and $z$ axis respectively and $(x,y,z)$ is the point at which we compute the convolution.

\subsection{U-Net}
\label{sec:unet}
\begin{figure}[ht]
\centering
\includegraphics[scale=.4]{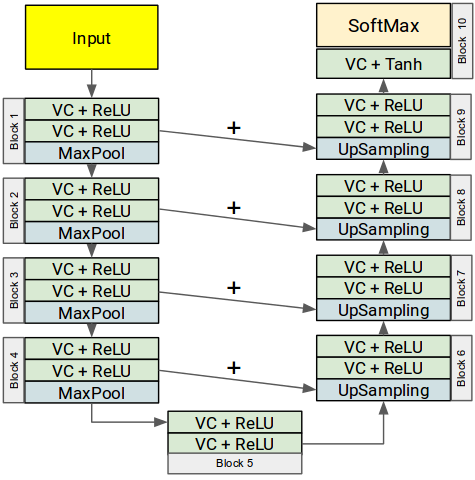}
\caption{U-Net Architecture. VC - Volumetric Convolution, ReLU - Rectified Linear Unit. For detailed hyperparameters of layers see Table~\ref{unettable}.}
\label{fig:unet}
\end{figure}

Up or fully convolutional neural networks (U-Nets) are a class of convolutional networks specifically designed for segmentation problems that were introduced by~\cite{ronneberger2015u}.
The network takes a spacial input and produces a condensed feature vector, by applying 3D convolutional and max pooling layers.
However, unlike traditional convolutional neural networks, after each down convolutional layer the output is remembered in order to mimic the down convolutional process in reverse. It does this by using the correspondingly sized output of the down step at each "up" step to construct a finer segmentation of the input, as seen in Figure~\ref{fig:unet}.

We train our model using Adam for automatic learning rate adjustment~\cite{kingma2014adam} and categorical cross entropy for the loss.
We chose to use rectified linear units (ReLU) as our activation function for all but the final two layers, in which we use hyperbolic tangent and a voxel-wise softmax layer respectively.
Experiments with hyperbolic tangent before softmax have consistently resulted in higher values of metrics and better segmentation.
Also batch normalization~\cite{ioffe2015batch} didn't improve the values metrics.
Further we will compare our model with best trained U-Net.
For a detailed description of layer sizes and the total number of parameters see Table~\ref{unettable}.
\begin{table}[ht]
\centering
\caption{Detailed hyperparameters for U-Net. $k^3$ means $k\times
  k\times k$. $M$ and $N$ is number of input volumes and number of
  classes respectively. 
  Because we are using just T1 MRI volume $M$ equal to 1 and segmenting gray matter, white matter and all outside the brain $N$ equal to 3.}
\label{unettable}
\begin{tabular}{|c|c|c|c|c|}
\hline
Block & \begin{tabular}[c]{@{}c@{}}Kernel\end{tabular} & \begin{tabular}[c]{@{}c@{}}Input\\ Feature\\ Maps\end{tabular} & \begin{tabular}[c]{@{}c@{}}Output\\ Feature\\ Maps\end{tabular} & \begin{tabular}[c]{@{}c@{}}MaxPool\\ UpSample\\ Kernel\end{tabular} \\ \hline
1     & $3^3$                                                 & $M$                                                            & $32$                                                            & $2^3$                                                                      \\ \hline
2     & $3^3$                                                 & $32$                                                           & $64$                                                            & $2^3$                                                                      \\ \hline
3     & $3^3$                                                 & $64$                                                           & $128$                                                           & $2^3$                                                                      \\ \hline
4     & $3^3$                                                 & $128$                                                          & $256$                                                           & $2^3$                                                                      \\ \hline
5     & $3^3$                                                 & $256$                                                          & $512$                                                           & ---                                                                    \\ \hline
6     & $3^3$                                                 & $512$                                                          & $256$                                                           & $2^3$                                                                      \\ \hline
7     & $3^3$                                                 & $256$                                                          & $128$                                                           & $2^3$                                                                      \\ \hline
8     & $3^3$                                                 & $128$                                                          & $64$                                                            & $2^3$                                                                      \\ \hline
9     & $3^3$                                                 & $64$                                                           & $32$                                                            & $2^3$                                                                      \\ \hline
10    & $1^3$                                                 & $32$                                                           & $N$                                                             & ---    \\ \hline
\multicolumn{3}{|c|}{Overall number of parameters}                                                                             & \multicolumn{2}{c|}{23 532 355}                                                                                                              \\ \hline
\end{tabular}
\end{table}

\subsection{MeshNet}
\label{sec:meshnet}
Our model is inspired by~\cite{yu2015multi}, where authors used CNN based context module with dilated kernels to improve dense prediction of the front-end module, which, in their case, was VGG-16~\cite{SimonyanZ14a} network.
Our architecture entirely consists of what~\cite{yu2015multi} used as a context module but we modified it to use 3D dilated convolutions (which we've implemented in~\cite{collobert2011torch7}).
We have used batch normalization~\cite{ioffe2015batch} which lead to a faster convergence.
Adam optimizer~\cite{kingma2014adam} and categorical cross-entropy loss as for U-Net.

Kernels of volumetric convolutions are necessarily dense because they need to span larger volume than in the 2D case.
This generates an increased number of parameters as they grow linearly with the number of neurons and layers.
For traditional CNN, for the neurons of the top layer to be able to project their receptive field onto a large extent of the input volume, one either needs to increase the depth of the network considerably, or increase the sizes of convolution kernels.
Dilated convolutions enable CNNs to flexibly control extent of the receptive field at the input while keeping the number of parameters fixed.

Discrete volumetric L-Dilated convolution can be expressed as follows:
\begin{equation}
(k*_lf)_{(x,y,z)}\!=\!\sum_{\bar{x}=-a}^a \sum_{\bar{y}=-b}^b \sum_{\bar{z}=-c}^c k(\bar{x},\bar{y},\bar{z})f(x-l\bar{x},y-l\bar{y},z-l\bar{z}),
\label{eq:dil3dconv}
\end{equation}
where $l$ is dilation factor, $*_l$ is $l$-dilated convolution.

Besides the flexible extension of the receptive field without paying a toll of increase of the parameter space, the main idea of the approach is for all layers to always operate in the same dimensions as the input.
The former allows flexible integration of information from multiple contexts.
The latter matches well the condition that the output labels occupy the same space as the input data and does not lose the information in transfer.
\begin{figure}[ht]
\centering
\includegraphics[scale=.4]{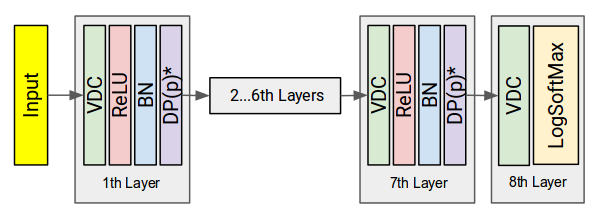}
\caption{MeshNet Architecture. VDC - Volumetric Dilated Convolution, ReLU - Rectified Linear Unit, BN - Batch Normalization, DP(p) - Dropout with probability $p$. For detailed hyperparameters of layers please look Table \ref{meshnettable}.}
\label{fig:meshnet}
\end{figure}

The architecture of the network is now a simple feed forward CNN (see Figure~\ref{fig:meshnet}).
This simplicity leads to minimization of model parameters as Table~\ref{meshnettable} shows.
Notably, for a model of better performance than U-Net (as we will show later), MeshNet has a parameter set more than 300 times smaller than that of U-Net.

\subsection{Metrics}
To measure the resulting performance of the model we are using DICE coefficient~\cite{taha2015metrics} and Average Volume Difference~\cite{mendrik2015mrbrains} for each tissue class: gray matter, white matter and all outside of the brain tissue (as background). 

DICE coefficient determines the spatial overlap of segmentation and it is defined as 
\begin{equation} \label{eq:dice}
  DICE = \frac{2 |A \cap G|}{|A| + |G|} = \frac{2TP}{2TP + FN + FP},
\end{equation}
where $A$ is segmentation by model and $G$ is the ground truth. Also DICE can be expressed as F1-Score. F1-score is $F_\beta$-score with $\beta = 1$ which is defined as
\begin{equation} \label{eq:fb_score}
F_{\beta}=\frac{(\beta^{2} +1)\cdot Precision \cdot Recall}{\beta^{2} \cdot Precision+Recall},
\end{equation}
where Precision and Recall is defined by equation \ref{eq:prec} and \ref{eq:recall} respectively, TP --- true positive, FP --- false positive, FN --- false negative.
\begin{equation} \label{eq:prec} Precision=\frac{TP}{TP+FP} \end{equation}
\begin{equation} \label{eq:recall} Recall=\frac{TP}{TP+FN} \end{equation}
And if we substitute equations \ref{eq:prec}, \ref{eq:recall} and $\beta=1$ to $F_\beta$ equation \ref{eq:fb_score} we will get F1-score equal to DICE coefficient.

Average Volume Difference is volumetric measure. It is defined as
\begin{equation} \label{eq:avd}
  AVD = \frac{|V_p \cap V_g|}{V_g},
\end{equation}
where $V_p$ is the volume of segmentation by model and $V_g$ is the volume of the ground truth for every tissue.

\begin{table}[ht]
\centering
\caption{Detailed hyperparameters for MeshNet with subvolume side length $A$ equal to 64 and 68. $k^3$ means $k\times k\times k$. $M$ and $N$ is number of input volumes and number of classes respectively. Because we are using just T1 MRI volume $M$ equal to 1 and segmenting gray matter, white matter and all outside the brain $N$ equal to 3.}
\label{meshnettable}
\begin{tabular}{|c|c|c|c|c|c|c|c|}
\hline
\multirow{2}{*}{Layer} & \multirow{2}{*}{Kernel} & \multirow{2}{*}{Input} & \multirow{2}{*}{Output} & \multicolumn{2}{c|}{A = 64}       & \multicolumn{2}{c|}{A=68}         \\ \cline{5-8} 
                       &                         &                        &                         & Pad             & Dil             & Pad             & Dil             \\ \hline
1                      & $3^3$       & $M \times A^3$         & $21 \times A^3$         & 1               & 1               & 1               & 1               \\ \hline
2                      & $3^3$       & $21 \times A^3$        & $21 \times A^3$         & 1               & 1               & 1               & 1               \\ \hline
3                      & $3^3$       & $21 \times A^3$        & $21 \times A^3$         & 1               & 1               & 2               & 2               \\ \hline
4                      & $3^3$       & $21 \times A^3$        & $21 \times A^3$         & 2               & 2               & 4               & 4               \\ \hline
5                      & $3^3$       & $21 \times A^3$        & $21 \times A^3$         & 4               & 4               & 8               & 8               \\ \hline
6                      & $3^3$       & $21 \times A^3$        & $21 \times A^3$         & 8               & 8              & 16              & 16              \\ \hline
7                      & $3^3$       & $21 \times A^3$        & $21 \times A^3$         & 1               & 1               & 1               & 1               \\ \hline
8                      & $1^3$       & $21 \times A^3$        & $N \times A^3$          & 0               & 1               & 0               & 1               \\ \hline
\multicolumn{4}{|c|}{Receptive Field}                                                               & \multicolumn{2}{c|}{$37^3$} & \multicolumn{2}{c|}{$67^3$} \\ \hline
\multicolumn{4}{|c|}{Overall number of parameters}                                                  & \multicolumn{4}{c|}{72516}                                            \\ \hline
\end{tabular}
\end{table}

\section{Results}
\label{sec:res}
\subsection{Datasets}

\subsubsection{Human Connectome} \label{human_connectome} Human Connectome~\cite{van2013wu} dataset contains close to a thousand subjects, each with a raw $256\times 256\times 256$ T1 volume and a processed FreeSurfer segmentation.
Using these segmentations, we didn't include subcortical and corpus callosum regions of Freesurfer in ground truth for training and testing sets.
We have randomly selected 20 subjects for training and a 100 other random subjects for testing.
Additionally we have used 2 other randomly selected subjects for validation of the models during training to control for potential overfitting.

\subsubsection{MRBrainS Challenge} MRBrainS Challenge dataset consists of twenty fully annotated multi-modal 3T MRI brain scans.
These scans are of subjects who are over fifty, have either diabetes or an increased cardiovascular risk, all of whom display varying degrees of atrophy and white matter lesions. For every subject T1, T1 IR and T2 FLAIR MRI images were provided.
All data is anisotropic with voxel size ($0.958mm \times 0.958mm \times 3.0mm$), bias corrected, aligned and has volume size $240 \times 240 \times 48$.
The dataset was divided by 5 and 15 subjects for training and test data respectively. In order to segment the brain, the manual labels were used as the ground truth and was only available during training.
Manual segmentation was done for 8 types of tissues: Cortical gray matter, Basal ganglia, White matter, White matter lesions, Cerebrospinal fluid in the extracerebral space, Ventricles, Cerebellum, Brainstem and everything else were labeled as background.
For every pair in the following list, they decided to assign the pair to its own individual class for testing purposes and everything else was left as the background class. The pairs were cerebrospinal fluid and ventricles, cortical gray matter and basal ganglia, white matter and white matter lesions.
For more information about dataset please look at http://mrbrains13.isi.uu.nl or article~\cite{mendrik2015mrbrains}.

\subsection{Volume sampling}
\begin{figure}[ht!]
\centering
\includegraphics[width=\linewidth]{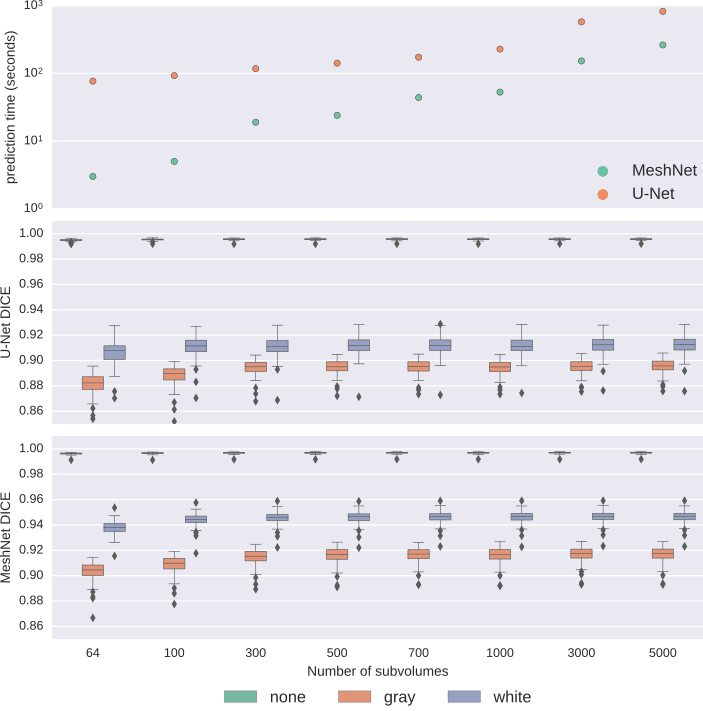}
\caption{Prediction time in logarithmic scale and accuracy measured as DICE coefficient relative to the FreeSurfer generated ground-truth as a function  of the  number of $64\times  64\times 64$ subvolumes used for prediction of T1 $256\times  256\times  256$ volumes for 100 testing brains previously unseen by the model. The number  of subvolumes that  were  used  for  covering  the   complete  T1  volume are $64, 100, 300, 500, 700, 1000, 3000, 5000$. For DICE coefficient higher is better and for AVD less is better.}
\label{fig:speed}
\end{figure}

\begin{figure}[ht]
\centering
\includegraphics[width=\linewidth]{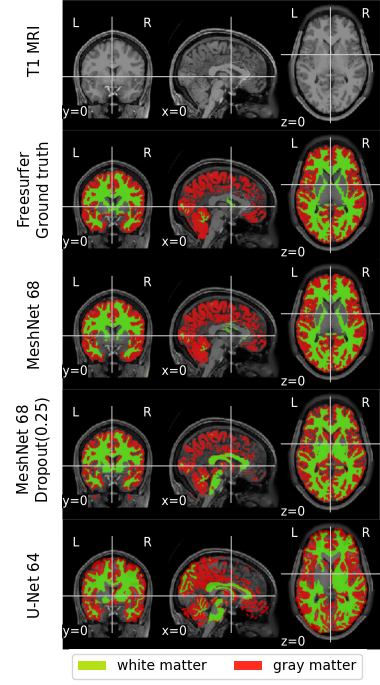}
\caption{An example of T1 MRI, Freesurfer segmentation (as ground truth) and prediction of MeshNet (with subvolume side length 68, without Dropout and  with probability 0.25 of Dropout) and U-Net (with subvolume side length 64).}
\label{fig:overlaps}
\end{figure}

\begin{figure}[ht]
\centering
\includegraphics[width=\linewidth]{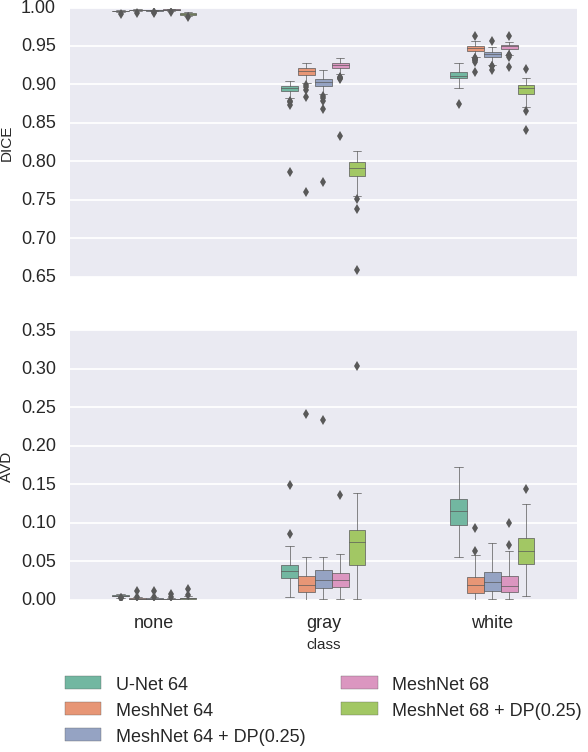}
\caption{DICE coefficient and AVD accuracy measures for predicting brain tissue types evaluated on three classes of tissues independently (gray matter, white matter, and outside the brain (none)) on 100 testing previously unseen by the model T1 $256\times 256\times 256$ volumes (each overlapped with 1000 subvolumes). Model name extensions 64 and 68 are for the side length of the input subvolume. For DICE coefficient higher is better and for AVD lower is better.}
\label{fig:metrics}
\end{figure}

For initial experiments on the Human Connectome dataset with isotropic $256\times256\times256$ volumes of T1 MRIs we used hyperparameters inherited from the 2D model of dilated network.
Unfortunately for $64\times 64\times 64$ subvolumes as input we failed to obtain good results, because in this case the receptive field of $67\times67\times67$ aperture that the top layer of MeshNet project down to the input layer is too high for this side length.
To correct for this we reduced our receptive field (at the input layer as seen by the output) to $35\times35\times35$.
Further we call  this model MeshNet 64.
Then we expanded receptive field back to the side-length of 67 voxels, but started to use $68\times68\times68$ side length based subvolumes as the input, which improved  result (as you can see for MeshNet 68 in Figure~\ref{fig:metrics}).
Thus, we recommend setting receptive fields close to the side length of the input subvolumes.

Randomly selected subvolumes represent a difficulty for training, since they quite often land in the majority of the volume, outside of the brain area, resulting in a highly imbalanced representation of classes and leading to the model simply preferring the background class.
To avoid this problem we use the following procedure.
We use a Gaussian distribution with mean $\vec{\mu}$ set at the center of the volume (usually corresponds with the center of the brain) and diagonal covariance $\Sigma$ with $\sigma_i = 50, i \in \{1,2,3\}$.
From this Gaussian, we sample points which represent the center of subvolumes which we then process with U-Net and MeshNet.
At training time, the process is random and indefinitely continuous.
This way we guarantee that rarely (if at all) does a network see the same training subvolume more than once.

For classifying full brain volumes using a model that classifies subvolumes we employ a method that will make sure everything is classified, but will also focus on where we care about most.
We start by segmenting the initial volume into a regular grid of subvolumes partitioning the whole volume.
These volumes ensure a prediction for each voxel.
Then we sample overlapping volumes from the brain region just like in the training, but here we stop sampling when we have $N$ subvolumes.
We then feed both models these same subvolumes to classify.
For each model we then stitch together a final predicted volume. 
This final volume's value at any voxel is chosen by a the majority vote of all the subvolume predictions that that voxel falls into.

\subsection{Performance comparison}

\begin{figure}[ht!]
\centering
\includegraphics[width=0.9\linewidth]{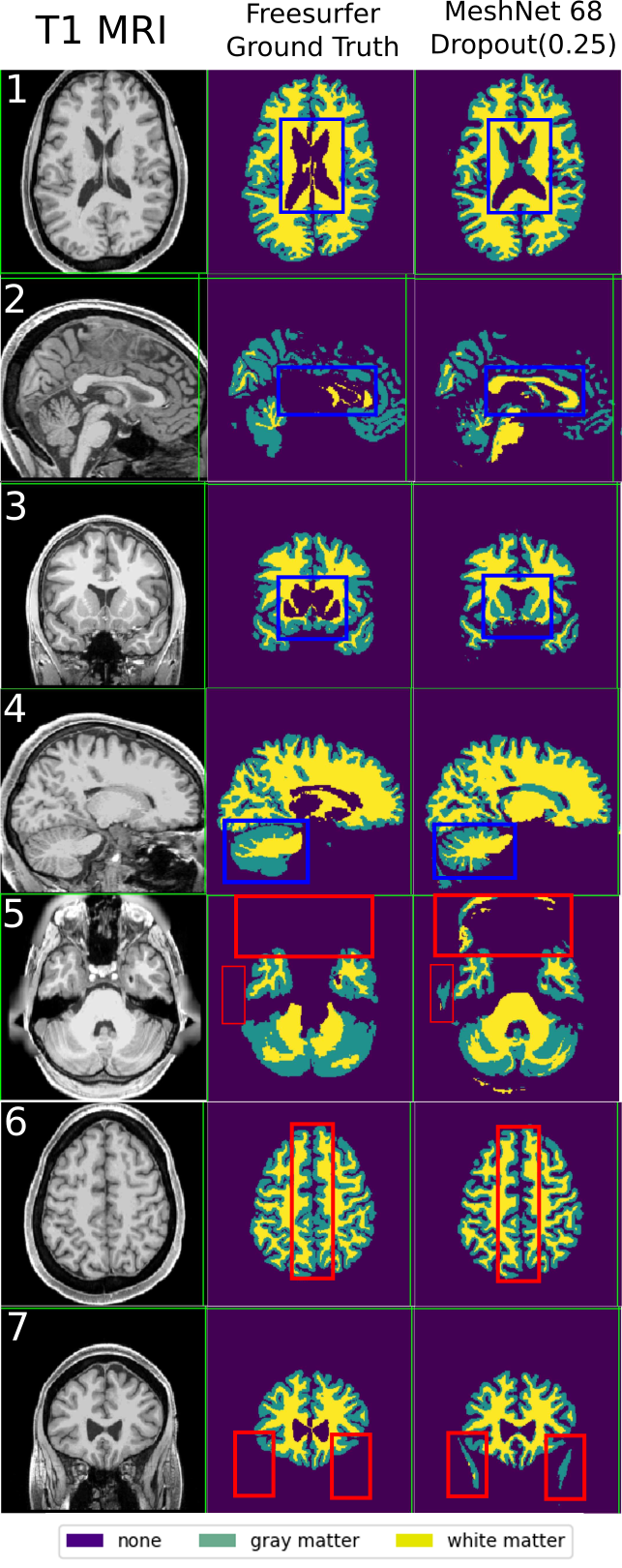}
\caption{Seven different slices with some good (Blue) and bad (Red) rectangles. Good means that our model performs better and bad --- when FreeSurfer is  better.}
\label{fig:pos_neg_examples}
\end{figure}

In order to compare the two deep learning architectures above, we trained each of them using 20 training volumes that had been classified by FreeSurfer.
These volumes underwent no preprocessing steps besides being normalized down to the unit interval.
The U-Net was trained for 2000 mini-batches of 64 cubes (seeing in total 128,000 subvolumes), and the MeshNet was trained on the same number of subvolumes as well.
We ran experiments to see how good our resulting segmentations were with different numbers of sampled volumes, see Figure~\ref{fig:speed}, to determine an approximate $N$ that gets some of the best accuracy the model has to offer with the least time, eventually settling on $N = 1000$. Note, however, that MeshNet attains an acceptable accuracy already at the $N=500$, where it runs in just little under 30 seconds. The time is measured on GeForce Titan X.
Compare to the approximately 2 minute segmentations time, made by the fastest models from the MBRainS Challenge on a smaller dataset, as well as our timing results on U-Net, which also ran in about 2 minutes.
We also have run experiments with 7000, 10000 and 20000 subvolumes for U-Net and MeshNet but DICE saturates and doesn't increase beyond the demonstrated values preserving the gap in the computational time and metrics between two methods.

Next we test and compare the models on 100 testing subjects' MRIs.
Figure~\ref{fig:metrics} summarizes the performance using DICE and AVD metrics.
As already expected, MeshNet without dropout better captures FreeSurfer labeling. 
Note, this is not overfitting in the classical machine learning definition, as we test on 100 previously unseen subjects, after training on only 20.
Rather this is trusting the teacher too much and, as we will see next, limiting the ability to generalize beyond imperfect labeling.
MeshNet with dropout shows the worst metrics of correspondence to FreeSurfer, while U-Net is in between.

However, the metrics can not tell the whole story on their own.
If the model somehow perfectly classified all testing volumes so that DICE coefficient is 1 and the AVD is 0 this would not be necessarily desirable.
We have to remember that our ground truth is produced by Freesurfer, a classification program that is good, but far from perfect~\cite{taha2015metrics}.
So a little imperfection in the gray and white matter classification relative this ``ground truth'' is perhaps desirable.
In order to get a better idea of what models are doing well we must look at the outputs and use that information along with metrics to make our choice.

\begin{figure}[ht]
\centering
\includegraphics[width=\linewidth]{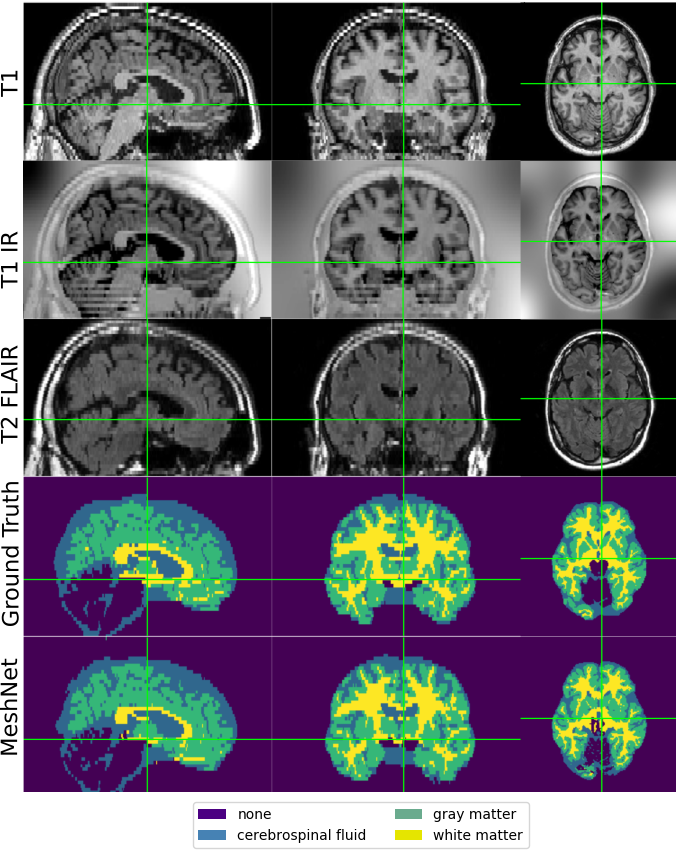}
\caption{An example of $240 \times 240 \times 48$ T1, T1 IR, T2 FLAIR MRI volumes, manual segmentation as ground truth provided by MRBrains Challenge~\cite{mendrik2015mrbrains} and our prediction of MeshNet with 1000 subvolumes $68\times68\times48$ without Dropout.}
\label{fig:MRBrains_example}
\end{figure}

\begin{figure}[ht]
\centering
\includegraphics[width=\linewidth]{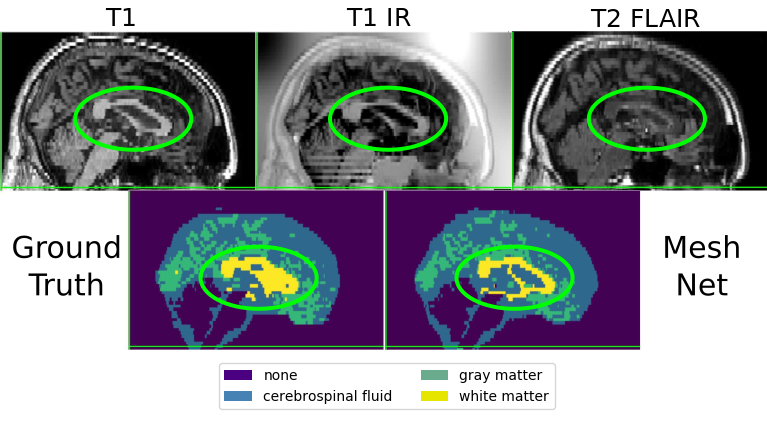}
\caption{The slice of one MRBrains Challenge volume in which our model performs well despite the error in ground truth.}
\label{fig:MRBrains_wrong}
\end{figure}

The results of classification by both of the methods, as well as the ``ground truth'' produced by FreeSurfer, are exemplified in Figure~\ref{fig:overlaps} which serves as an example of how each method performs.
The following conclusions about performance of the models are made by experts in brain imaging and Freesurfer processing after they had checked the results of segmentation.
Upon inspection we see that MeshNet without dropout seems to look very similar to Freesurfer. Conversely, MeshNet with dropout seems to improve upon the previous one by classifying white matter near the center of the brain that the previous two ignored, showing that it is learning how to classify volumes in a certain generality that it did not learn without dropout.
Note, as we mentioned in Section~\ref{human_connectome}, we didn't include subcortical and corpus callosum in the ground truth.
When we consider the U-Net output, it becomes clear that it is asymmetric, classifying central white matter not in the Freesurfer classification on the right lobe, but not on the left lobe.
While it is entirely possible that with the correct tweaks the U-Net model may be able to preform just as well as the MeshNet model, in light of the significant time difference between the classification times of the two models and the obvious success of the MeshNet with dropout, for the rest of this paper we will focus on the MeshNet model.

Having chosen a fast end to end model with good statistical properties which has demonstrated an ability to learn in generality, we now wish to see just what general facts it has learned.
In Figure~\ref{fig:pos_neg_examples} we compare MeshNet and Freesurfer classifications of a volume.
Blue rectangles highlight cases where MeshNet outperforms the teacher (FreeSurfer), while red point to its mistakes.
In the top half of the figure it is fairly easy to see that MeshNet does a more delicate and natural job when classifying white and gray matter, particularly near the center of the volume.
However, below that we see where it did worse than Freesurfer.
For example, MeshNet occasionally will mis-classify skull and neck regions as brain regions.
This is a problem that has potential fixes through some pre or post processing of the volumes.
Unfortunately, MeshNet also seems to advocate for a volume that is more separated between the left and right lobes than would be desirable.
All together though, MeshNet seems to have surpassed its teacher in many useful areas of classification.

Combining two models---MeshNet with and without dropout---in this case does not guarantee an improved result.
The majority voting scheme will likely result in the average of two volume-sampled predictions or we again will use imperfect labeling of Freesurfer.
MeshNet without dropout tends to be worse in the center of brain, but decreases gap between side lobes, mistakes in the neck and skull regions by MeshNet with dropout.

\subsection{Expert labeling}
As we already mentioned, we have used FreeSurfer segmentation as the ground truth, but it can not be used as a gold standard in the way manual segmentation is.
To check performance of our model on manually segmented brains we have trained MeshNet on a dataset with 5 brains from the MRBrainS Challenge.
We used 3 subjects to train the model, 1 subject to validate/avoid overfitting and 1 subject to test. The testing result is shown on Figure~\ref{fig:MRBrains_example}.
The metrics that our model scored on this volume are as follows, DICE: background $99.4\%$, csf $81.9\%$, gray matter $87.7\%$, white matter $89.9\%$ and AVD: background $0.3\%$, csf $6.8\%$, gray matter $3.4\%$, white matter $6.6\%$.
With these results our model can get into the top 10 ranking of MRBrainS Challenge.\footnote{Exact ranking isn't available due to submission process.}
However, manual segmentations can also have mistakes in the labeling, which Figure~\ref{fig:MRBrains_wrong} clearly points out.
Our approach can still train and perform well despite mislabelings in the ground truth.
The average time of running MeshNet on GeForce Titan X (Pascal) is 33.6 seconds, as calculated on 15 test subjects with 1000 subvolumes.

Another finding in this experiment is the ease with which the model could integrate more than a single input. Potentially, we could improve the models performance if we would use other modalities whose output data is confined to the same volume space (e.g .diffusion tensor imaging (DTI)).

\section{Conclusions}
\label{sec:conc}
With the growing volumes of biomedical data it becomes increasingly important to be able to automate and speed up the processing of the collected data.
In this paper we have presented a method for brain tissue segmentation in MRI images from unprocessed input data.
Our approach is faster than the state of the art methods according to MBRainS ranking table and our comparisons with U-Net, requires fewer data to train due to its reduced parameter set and is able to generalize from imperfect training data.
All of these properties make it an attractive tool, as it can be trained on the data that is already available thanks to existing segmentation approaches, and yet it can outperform the methods used to train it.
Furthermore, in some cases the proposed method can even outperform the expert labeling.
All of the above can make the method useful for research and clinical applications, possibly beyond brain imaging.
With further work MeshNet can be implemented more efficiently and achieve even faster execution times.
In neither of our experiments did we use data augmentation, we only trained on what's available in the raw input.
We can potentially improve the performance by using the standard data augmentation techniques.

\bibliographystyle{IEEEtran}
\bibliography{references}
\end{document}